\documentclass[runningheads,a4paper]{llncs}

\usepackage{amssymb}
\usepackage{amsmath}
\usepackage{graphicx}

\usepackage{url}

\usepackage{xcolor}
\usepackage{amsmath}
\usepackage{xspace}

\newcommand{\mi}[1]{\text{\textit{#1}}}

\newcommand{\keywords}[1]{\par\addvspace\baselineskip
\noindent\keywordname\enspace\ignorespaces#1}
\newcommand{\clpqs}{CLP(QS)\xspace}
\begin{document}

\mainmatter

\title{Grounding Dynamic Spatial Relations for Embodied (Robot) Interaction}


\titlerunning{Grounding Dynamic Spatial Relations}

\author{Michael Spranger$^1$ \and Jakob Suchan$^2$ \and Mehul Bhatt$^2$ \and Manfred Eppe$^3$}
\authorrunning{Suchan, Spranger, Bhatt, Eppe} 

\institute{
$^1$Sony CSL,\\
 3-14-13 Higashi Gotanda,\\
141-0022 Tokyo \\
$~$\\
$^2$Cognitive Systems, and\\
Spatial Cognition Research Center (SFB/TR 8),\\
University of Bremen, Germany\\
$~$\\
$^3$ IIIA-CSIC,
Spain}

%



\maketitle

\begin{abstract}
This paper presents a computational model of the processing of dynamic spatial relations occurring in an embodied robotic interaction setup. A complete system is introduced that allows autonomous robots to produce and interpret dynamic spatial phrases (in English) given an environment of moving objects. The model unites two separate research strands: computational cognitive semantics and  on commonsense spatial representation and reasoning. The model for the first time demonstrates an integration of these different strands.

\end{abstract}

\section{Introduction}
Commonsense spatio-linguistic abstractions offer a human-centred and cognitively adequate mechanism for the computational handling of spatio-temporal information in a wide-range of cognitive interaction systems, and cognitive assistive technologies \cite{Bhatt-Schultz-Freksa:2013}. Recently, there has been extensive work on static spatial relations such as ``front'', ``back'', ``left'' etc. We now have working models for the processing of static spatial relations \cite{moratz2006spatial,kelleher2006proximity}, the learning of spatial phrases \cite{regier2005emergence}, and their evolution \cite{spranger2012irl}. At the same time, formalizations (e.g., logical, relational-algebraic) of space and development of tools for efficiently reasoning with spatial information is a vibrant research area within knowledge representation and reasoning (KR) \cite{bhatt2011-scc-trends,renz-nebel-hdbk07}. Commonsense spatial and temporal representations abstract from exact numerical representations by describing relations between objects using a finite set of relations. Spatial change is often modelled using \emph{conceptual neighborhoods} \cite{Freksa1991}. Relations between two entities are conceptual neighbors if they can directly be transformed from one relation to another by continuous change of the environment. Researchers have investigated movement on the basis of an integrated theory of space, time, objects, and position \cite{Galton2000} or defined continuous change using 4-dimensional regions in space-time \cite{Muller1998}. Davis \cite{Davis2012} discusses the use of transition graphs for reasoning about continuous spatial change and applies them in physical reasoning problems. Qualitative spatial calculi have been integrated with the \emph{situation calculus} family of action logics \cite{Bhatt:RSAC:2012,bhatt-et-al-2011}. A particular emphasis has been on the formal and computational characterisation of `space' and dynamic `visuo-spatial' problem-solving processes within a range of spatial assistance systems involving space and language \cite{Bhatt-Schultz-Freksa:2013}.

On the other hand, there has been extensive work on modeling spatial relations for language. Models of proximal relations \cite{kelleher2006proximity} based on proximity fields, for projective and absolute spatial relations based on prototypes \cite{moratz2006spatial} and group-based reference \cite{tenbrink2007space} have been proposed. Static spatial relations are interesting but they only cover a particular subdomain of spatial relations namely relations that do not encode temporal qualities. Recent models of dynamic spatial relations using semantic fields \cite{fasola2013using} or probabilistic graphical models \cite{tellex2011approaching} try to deal with temporal aspects of spatial relations. However, none of these models have attempted to integrate qualitative spatial reasoning systems which are able to deal with missing or unobserved data. This is somewhat of a surprise since the spatial reasoning community has provided sophisticated tools, frameworks and theories for modeling both static and dynamic spatial relations \cite{bhatt2011-scc-trends,Bhatt:RSAC:2012}. 

This paper reports progress in combining the work on reasoning with cognitive semantics models of dynamic spatial relations. The focus of this paper is a model of dynamic spatial relations such as ``across'', ``over'', ``into'', ``out off'' by integrating methods from cognitive computational semantics with commonsense spatial representation and reasoning techniques. The system presented here is capable of processing phrases such as ``the yellow block \emph{moves across} the red region''. The following sections will introduce the interaction scenario, followed by a more detailed description of the modules comprising the model.

\section{Embodied Interaction}
\label{s:setup}

\begin{figure}[t]
\begin{center}
\includegraphics[width=1.0\columnwidth]{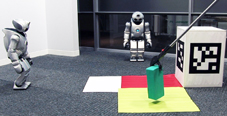}
\end{center}
\caption{{\sffamily Spatial setup. Objects move in an environment that features landmarks such as the box, and regions (colour patches on the ground). To the left the scene model extracted by the left robot is shown, to the right the scene model computed by the right robot is shown. 
The estimated movement of the block is signified through opacity. The starting point has a lower opacity (alpha) than the finish point. Regions are signified by the coloured rectangles. The blue square is the estimate of position and orientation of the box. Arrows signify robots. The lower centre arrow in each model stands for the own position (origin of the coordinate system). The other arrow visualises the estimated position and orientation of the other robot.}}
\label{f:setup}
\end{figure}

One of the key methodologies for research and validation of 
theories of grounding are situated interactions called
\emph{language games} \cite{steels2003evolving}. Language games are interactions
of two or more agents in the real (or in a simulated) world. 
Figure \ref{f:setup} shows the environment 
in which two robots interact. For the experiments presented in this paper, we used \emph{Sony 
humanoid} robots.  Robots are equipped with a
vision system that fuses information from the robot's camera with
proprioceptive sensors distributed across the body. The vision system 
singles out and tracks objects \cite{spranger2012perception}.
Here, the environment contains four types of objects: 
\emph{blocks}, \emph{boxes}, \emph{robots}. The vision system extracts the objects (as blobs) from the
environment and computes a number of raw, continuous-valued
features such as \emph{x}, \emph{y} position, \emph{width}, and 
\emph{height} and colour values (YCbCr). Objects are tracked over time and assigned
unique identifiers. So for instance, the green block has been given
the arbitrary id {\small\tt obj-12} by the left robot. The same robot identifies the white region as {\small\tt reg-36}.
The identifier stays the same for the period of time, where the robot is able to establish spatial-temporal
continuity. We recorded a number of scenes with varying spatial combinations of objects, regions and landmarks.

A spatial language game follows a script between
two randomly drawn agents from the population $P$ of agents.
One acts as the speaker, the other as the hearer. The agents see
two scenes in succession that differ in terms of the movement of 
objects (but typically not in the number, type or colour of objects).
For instance, in scene one an object might move from a yellow region
to a white region over the red region. In the second scene, a similar
object might move from the yellow region to the white region but without
moving across the red region (see 
Fig. \ref{f:setup} as example).

\begin{enumerate}{\small
\item The robots perceive two scenes and construe qualitative relations for each scene.
\item The speaker selects a scene from the two observed scenes, called the topic scene $T$.
\item The speaker \emph{conceptualizes} a meaning comprised of 
dynamic spatial relations, and construal operations for discriminating $T$.
E.g. the speaker computes that $T$ is different from the second scene in that the 
yellow block crosses a red region.
\item The speaker tries to express the conceptualization using an English grammar. 
For instance, the robot might \emph{produce} ``the green block moves across the red region''.
\item The hearer \emph{parses} the phrase using his English grammar, thereby computing the meaning underlying the phrase.
\item When the hearer was able to parse the phrase or parts of the phrase,
he examines the two scenes to find the scene which satisfies the conceptualization.
strategy (\emph{interpretation}).
\item The hearer signals to the speaker which scene he thinks the sentence is about\footnote{Signalling is done by using extra-linguistic feedback. Two gestures are assigned the meaning scene-1 or scene-2. The knowledge which gesture refers to which scene is known to all participant agents.}.
\item The speaker checks whether the hearer selected the same 
scene as the one he had originally chosen. If they are the same, 
the game is a \emph{success} and the 
speaker signals this outcome to the hearer. Otherwise, the game is a \emph{failure}. In that case the speaker signals the topic 
$T$ scene.}
\end{enumerate}

\begin{figure}[t]
\begin{center}
\includegraphics[width=0.75\columnwidth]{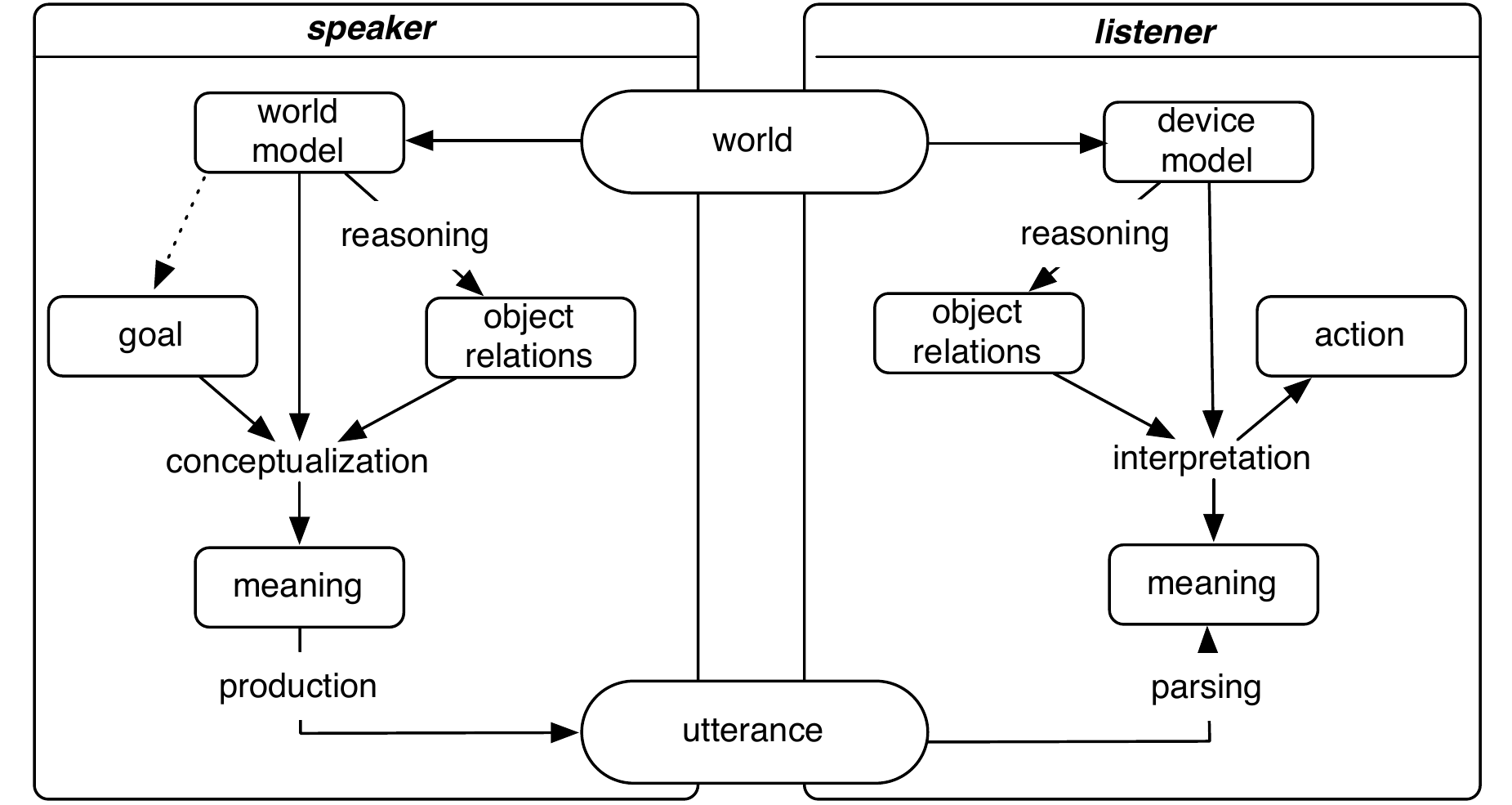}
\end{center}
\caption{{\sffamily Schematic of the systems involved in autonomous processing
of spatial scenes such as the one shown in Figure \ref{f:setup}.
Left are the processes the speaker runs, in order, to produce an utterance. 
Right shows the processes the hearer is running for understanding a phrase.}}
\label{f:semiotic-cycle}
\end{figure}

Agents are equipped with a range of software systems that allow 
them to participate successfully in interactions.
The following sections detail the main components of the system 1) a qualitative spatial reasoner
that provides qualitative descriptions of dynamic scenes,  2) a cognitive semantics system 
that picks distinctive qualitative aspects of a dynamic scene in order to discriminate between
scenes and 3) a construction grammar implementation of dynamic spatial phrase processing
that allows to produce and parse correct English sentences. See Figure \ref{f:semiotic-cycle}
for an overview of processing steps involved.

\begin{table}[t]
\centering 
\footnotesize
  \renewcommand{\arraystretch}{1.25}
\begin{tabular}{ |l c| }
  \hline
  \multicolumn{2}{|c|}{\textbf{\normalsize $\Sigma$ Space}} \\
  \hline
  \multicolumn{2}{|l|}{\textbf{Topology} }\\
  \multicolumn{2}{|c|}{$\mi{DC}(p,q,t), \mi{PO}(p,q,t), \mi{PP}(p,q,t), \mi{PPi}(p,q,t)$, $\mi{EQ}(p,q,t)$} \\[3pt]
  \multicolumn{2}{|l|}{\textbf{Extrinsic Orientation} (horizontal and in depth) }\\
  \multicolumn{2}{|c|}{ $\mi{left}(p,q,t), \mi{overlaps\_left}(p,q,t), \mi{along\_left}(p,q,t), \mi{horizontally\_equal}(p,q,t), $}\\
  \multicolumn{2}{|c|}{ $\mi{overlaps\_right}(p,q,t), \mi{along\_right}(p,q,t), \mi{right}(p,q,t)$}\\[6pt]
  \multicolumn{2}{|c|}{ $~$ $\mi{closer}(p,q,t), \mi{overlaps\_closer}(p,q,t), \mi{along\_closer}(p,q,t), \mi{distance\_equal}(p,q,t), $ $~$ }\\
  \multicolumn{2}{|c|}{ $\mi{overlaps\_further}(p,q,t), \mi{along\_further}(p,q,t), \mi{further}(p,q,t)$}\\
  \hline
  \hline
  \multicolumn{2}{|c|}{\textbf{\normalsize $\Sigma$ Motion}} \\
  \hline
  \textbf{Movement}& $\mi{approaching}(p,q,t)$ and $\mi{receding}(p,q,t)$  \\
  \hline 
\end{tabular}
\vspace{10pt}
\caption{{\sffamily Spatial relations used to describe the spatial configuration of a scene and the corresponding motion relations.}}   
  \label{tab:space_motion}
\end{table}

\section{Reasoning for Dynamic Spatial Relations}

Robots compute qualitative representations of space and motion. From these representations, spatio-temporal relations holding between spatial entities (objects) in the environment follow, i.e. \emph{topology, orientation, movement}.  The theory is implemented on top of \clpqs \cite{bhatt-et-al-2011}, which is a \emph{declarative spatial reasoning framework} that can be used for representing and reasoning about high-level, qualitative spatial knowledge about the world.\footnote{CLP(QS): A Declarative Spatial Reasoning System. \url{www.spatial-reasoning.com} } \clpqs  implements the semantics of qualitative spatial relations within a constraint logic programming framework (amongst other things, this makes it possible to use spatial entities and relations between them as native entities). 
In the following we describe the qualitative abstractions of space and motion and the thereon defined object relations, which serve as a basis to generate and interpret descriptions of the scene used by the robots in the description game.  

\subsection{Qualitative Abstractions of Space and Motion}
Based on perceived objects and regions represented by numerical features, the robots compute qualitative abstractions of space and motion (see Table \ref{tab:space_motion}) \cite{ECCV-narrative-2014}. The qualitative abstractions are obtained from the sensory data by applying thresholds on the continuous feature values (position, size, angle) estimated from objects in the scene. The detected entities are represented by the basic domain primitives: \emph{regions, points}, and \emph{oriented-points}.

\subsection{$\Sigma$ Space -- Qualitative Spatial Relations} 
Spatial primitives are the basis for computing the spatial relations \emph{topology}, and \emph{extrinsic-orientation}. 

\begin{description}

\item \textbf{Topology} \quad 
We represent the connectedness of pairs of spatial primitives by the relations of the region connection calculus \cite{Cohn1997}, using the RCC5 \cite{Cohn1997} subset in a ternary version, where the third argument represents the time point at which the relation holds.

\medskip
\item \textbf{Extrinsic Orientation (Position)} \quad
We represent the extrinsic orientation (relative position) of two spatial primitives, with respect to the observer's viewpoint,
making distinctions on the \emph{position} and the \emph{size} of the spatial entities.  
For the described robot scenario, we use a 2-Dimensional representation that resemble Allen's
interval algebra \cite{Allen1983} for each dimension, i.e. \emph{vertical}, and \emph{depth} (distance from the observing robot). However, in terms of visual perception, the interval relations that happen ``instantaneously'' (namely, $meets$, $starts$, and $finishes$) are irrelevant.

\end{description}

\subsection{$\Sigma$ Motion -- Qualitative Spatial Change}

 The dynamics of scenes are represented in terms of the \emph{relative movement} of the objects \cite{ECCV-narrative-2014}.

\begin{description}

\item \textbf{Relative Movement}\quad
The relative movement of pairs of spatial primitives is represented in terms of changes in the distance between the \emph{centroids} of the entities.
We represent these changes in terms of $approaching$ and $receding$ as defined below.

\begin{subequations}
\footnotesize
\scriptsize
\begin{align}
& \mi{approaching}(p,q,t) \leftrightarrow \exists t_1 t_2 (t_1 < t)\wedge (t < t_2)\wedge (\mi{dist}(p,q, t_2) < \mi{dist}(p,q,t_1));\\[1pt]
& \mi{receding}(p,q,t) \leftrightarrow \exists t_1 t_2 (t_1 < t)\wedge (t < t_2)\wedge (\mi{dist}(p,q,t_2) > \mi{dist}(p,q,t_1)). 
\end{align}
\normalsize
\end{subequations}

Notice that the timepoint $t$ falls within the open time interval (t1, t2), which is assumed to be very small; therefore, these predicates are locally valid with respect to the time point \cite{ECCV-narrative-2014}.


\end{description}

\medskip

\begin{figure*}[t]
  \centering
  \includegraphics[width=1\textwidth]{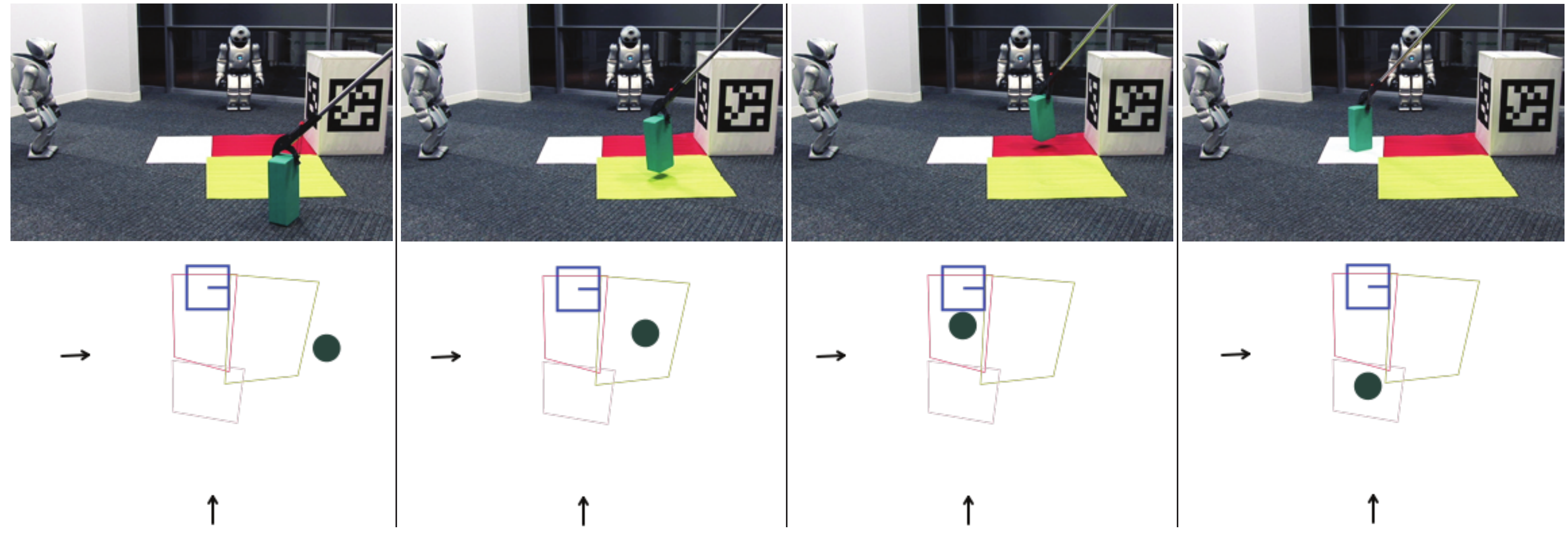}
      \caption{{\sffamily Dynamic scene as observed by the robot to the left.}}   
  \label{fig:dynamic_data}
\end{figure*}

\subsection{Spatio-Temporal Object Relations}

To describe perceived scenes in terms of spatio-temporal phenomena we combine the different aspects of the theory about \emph{space} and \emph{motion} providing a rich vocabulary about qualitative changes in the visual domain. This allows us to describe the ongoing interactions and operations between the physical entities represented by the spatial entities. 

\begin{description}
\item \textbf{Robots and objects} \quad Spatial scenes (for the purpose of this paper) consist of blocks, regions and robots. For the objects in the scene we assume that they are all ridged and non-opaque. For the regions and robots, we assume, that they are static. This means that only blocks are considered movable, so any observed change in the scene is due to movement of the blocks. Additionally, robots are assumed to have a certain orientation (facing direction) and we define abstract objects to represent the robots field of view.
\end{description}

Object relations are defined on the relations of space and motion and temporal relations (similar to Allen's Interval Algebra \cite{Allen1983}). These relations describe changes in the spatial configuration and motion of the objects and robots in the environment and are used by the robots to generate and interpret descriptions used in the discrimination game. E.g., a block B enters a region R at an time interval I.
%
%
%
\scriptsize
\begin{align*}
& \mi{moves\_into}(\mi{object}(B), \mi{region}(R), I) \leftarrow \mi{approaching}(\mi{object}(B), \mi{region}(R), I_4) \wedge \\
& ~~~DC(\mi{object}(B), \mi{region}(R), I_1) \wedge PO(\mi{object}(B), \mi{region}(R), I_2) \wedge PP(\mi{object}(B), \mi{region}(R), I_3) \wedge \\
& ~~~\mi{during}(I_2, I_4) \wedge \mi{equal}(I, I_2) \wedge \mi{meets}(I_1, I_2) \wedge \mi{meets}(I_2, I_3).
\end{align*}
\normalsize

%

A complex interaction as e.g. a block B moving across a region R is then defined based on the basic interactions, i.e., \emph{moves\_into, moves, moves\_out\_of} etc.  
\scriptsize
\begin{align*}
& \mi{moves\_across}(\mi{object}(B), \mi{region}(R), I) \leftarrow \mi{moves}(\mi{object}(B), \mi{direction}(Dir), I_1) \wedge \\
&  ~~~~~ \mi{moves\_into}(\mi{object}(B), \mi{region}(R), I_2) \wedge \mi{moves\_out\_of}(\mi{object}(B), \mi{region}(R), I_3) \wedge \\
&  ~~~~~   \mi{during}(I, I_1) \wedge \mi{starts}(I_2, I) \wedge \mi{finishes}(I_3, I)
\end{align*}
\normalsize


 \begin{figure*}[t]
  \centering
  \includegraphics[width=1\textwidth]{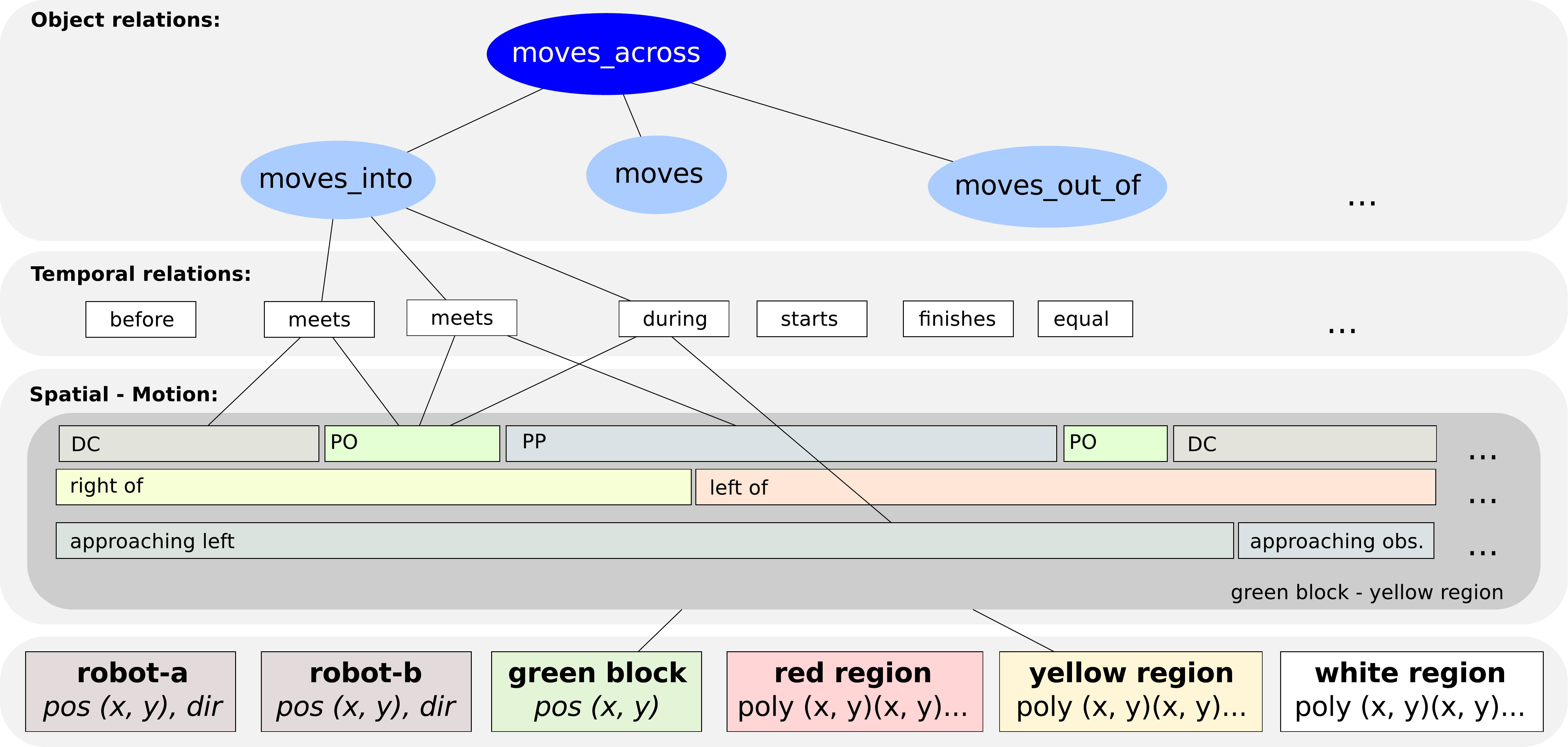}
      \caption{{\sffamily Object Relations based on qualitative abstractions of space and motion used to generate and interpret scene descriptions}}   
  \label{fig:spatio-temporal-obj-rel}
\end{figure*}

%
%
%
%
%

Based on these object relations a description of the scene is generated by means of declarative logic programming. An exemplary description of the scene shown in Fig. \ref{fig:dynamic_data} using the perspective of one of the robots (robot-a): 
\scriptsize
\begin{align*}
& \mi{moves}(\mi{object}(\mi{obj-12}), \mi{direction}(left), I_{1}) \wedge \mi{moves\_into}(\mi{object}(\mi{obj-12}), \mi{region}(\mi{reg-38}), I_{2}) \wedge \\
& \mi{moves\_across}(\mi{object}(\mi{obj-12}), \mi{region}(\mi{reg-38}), I_{3}) \wedge  \mi{moves\_into}(\mi{object}(\mi{obj-12}), \mi{region}(\mi{reg-37}), I_{4}) \wedge \\
& \mi{moves\_out\_of}(\mi{object}(\mi{obj-12}), \mi{region}(\mi{reg-38}), I_{5}) \wedge  \mi{moves}(\mi{object}(\mi{obj-12}), \mi{direction}(\mi{closer}), I_{6}) \wedge \\
& \mi{moves\_out\_of}(\mi{object}(\mi{obj-12}), \mi{region}(\mi{reg-37}), I_{7}) \wedge \mi{moves\_into}(\mi{object}(\mi{obj-12}), \mi{region}(\mi{reg-36}), I_{8})
\end{align*}
\normalsize

%
\noindent These object relations (as depicted in Fig. \ref{fig:spatio-temporal-obj-rel}) can be uttered with the following natural language description:

\begin{quote}
``The green block moves left, across the yellow region and enters the red region. It moves closer and leaves the red region at the bottom. The block enters the white region and stops within this region.''
\end{quote}

\section{Conceptualization and Interpretation}

\begin{figure}[t]
\begin{center}
\includegraphics[width=1\columnwidth]{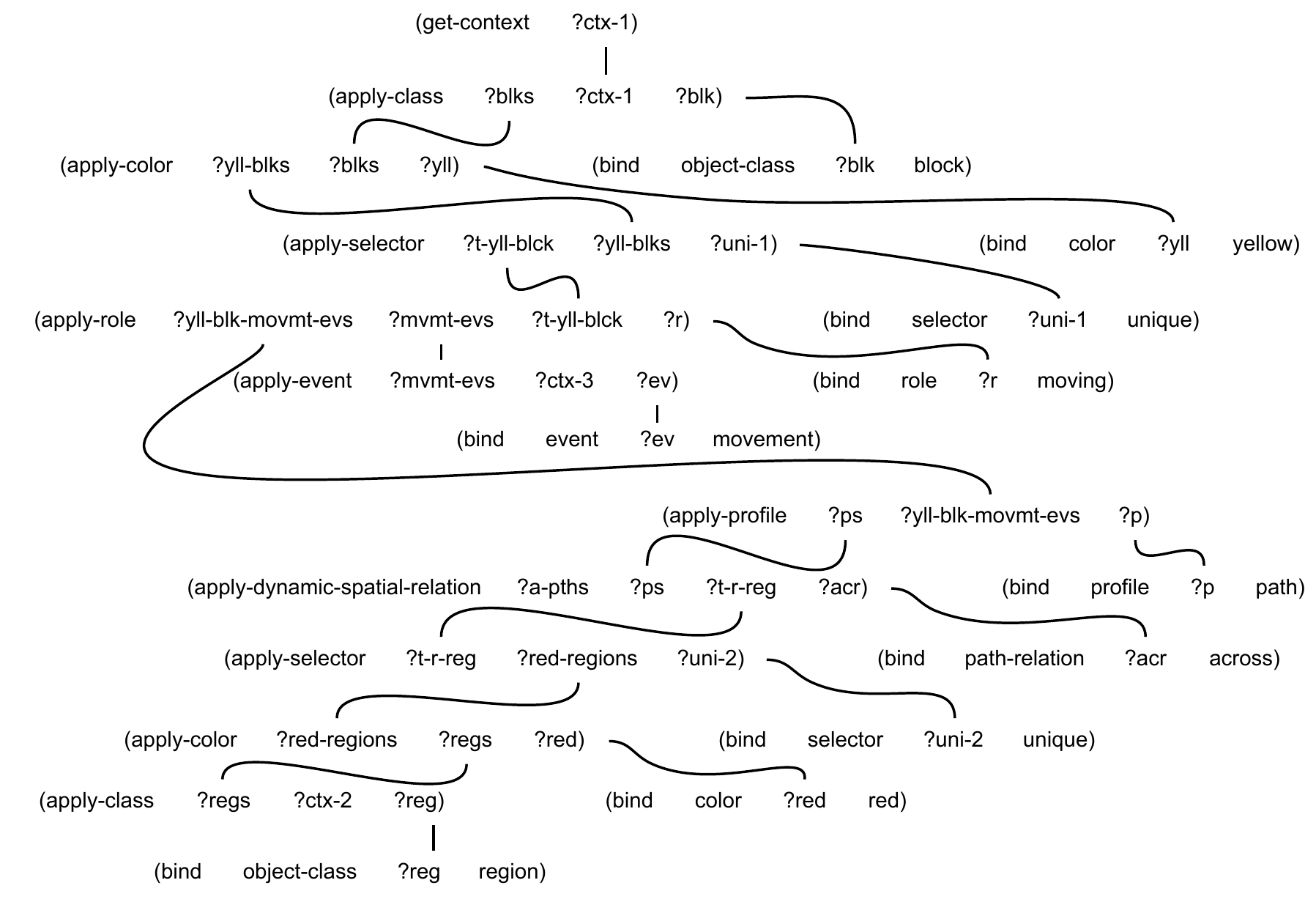}
\end{center}
\caption{{\sffamily IRL-program representing the semantic structure of the 
phrase ``the yellow block is moving across the red region''. The connections between
the operations and bind-statements signify variables
appearing in multiple places of the graph.}}
\label{f:irl-program}
\end{figure}

Robots use temporal object relations to pick a particular semantics 
discriminating or describing a scene. To represent the semantics of 
spatial phrases, we use a Cognitive semantics inspired formalism called Incremental Recruitment Language (IRL) 
\cite{spranger2012irl}.
The key idea behind this formalism is that the semantics 
of natural language can be modeled as a \emph{semantic program} 
\cite{johnson1977procedural}. In IRL, the meaning of an utterance 
consists of an algorithm and data pointers that when executed 
by the hearer will lead him to identify the topic (i.e. topic scene).

Let us exemplarily consider the phrase ``the yellow block moves across the red region''.
The phrase consists of a dynamic spatial relation (across), concepts such as box and region 
and determiners. The example phrase encodes a particular strategy for conceptualization 
where the spatial relation is used in conjunction with a region. An interpreter of the phrase
has to construe the path \cite{talmy2000toward1} of the object using the region and the particular
dynamic spatial relation.

Figure \ref{f:irl-program} shows a graphical representation
of the IRL-program underlying the phrase. Such programs consist
of two things and links between them.
\begin{description}
\item{\bf Cognitive operations} represent algorithms used in conceptualization. They encode a particular 
cognitive function such as categorization using a spatial category, 
applying a selector or applying an object class such as region or box.  
Cognitive operations are identified by their name, e.g. {\small\tt apply-class}
and they have a set of arguments, which can be linked to other 
operations or semantic entities via variables (starting with {\small\tt ?}).
\item{\bf Semantic entities} are the data that cognitive operations work with. 
They can be prototypes, concepts and categories or more generally representations 
 of the current context, as well as data exchanged between cognitive operations. 
 They are introduced explicitly in the network via \emph{bind-statements}. 
 For instance, the statement  {\small\tt (bind dynamic-spatial-relation ?acr across)}
 encodes the access to the agent-internal, dynamic spatial relation {\small\tt across} which
 will be bound to the variable {\small\tt ?acr}.
\end{description}

\subsection{Evaluation}

Evaluation is a process by which IRL-programs get executed. The process determines bindings for 
the variables in an IRL-program given a particular dynamic spatial scene or a number of spatial scenes. 
This process can fail, for instance, when a particular spatial scene does not fit the program. More precisely,
evaluation can succeed, or fail, but all successful evaluations are also scored \cite{spranger2012deviation}, as to how much the program
fits the scene.

Let us assume the hearer wants to interpret the example phrase and has decoded the IRL-program 
in Figure \ref{f:irl-program}. Evaluation proceeds as follows. First all 
bind-statements are evaluated, after which, for example, the variable {\small\tt ?blk} is bound 
to the concept {\small\tt block} and {\small\tt ?acr} to the spatial category {\small\tt across}. 
After that, the evaluation engine will try and  find cognitive operations that can be evaluated. Probably, the first cognitive operation 
to be evaluated is {\small\tt get-context} which binds the variable {\small\tt ?ctx-1} to each of the
2 scenes presented to the agent. Next {\small\tt apply-class} identifies blocks and regions 
from the context. After that {\small\tt apply-color} filters regions and blocks using colors.
Then {\small\tt apply-determiner} applies uniqueness constraints. 

The cognitive operations are implemented by integrating ideas from prototype theory and spatial reasoning. The following gives a brief and incomplete overview of the inner workings of the most important operations%
\footnote{This is a simplified description of the actual algorithm, cognitive operations implement multiple input/output patterns}
\begin{description}
\item{{\small\tt apply-event}} (for our purposes here) applies the movement event descriptor and computes a set of events/trajectories that can be categorized as movements. For the scene discussed in Section \ref{s:setup}, one movement event will be identified. The representation of these event includes information of the type (movement), event participants involved in the event (green block), as well as the time frame for it. Notice that at this stage no qualitative spatial information is used. There is a threshold for what is considered a movement vs. a non-movement and a classifier identifies those timeframes and objects that are considered part movements. For this paper only movement events are considered but the same mechanisms extend to complex events such as grasping, pushing etc.
\item{{\small\tt apply-role}} filters events for their participants. Here, the green block has to be in the moving role of the event. All movement events which are not involving a moving green block are filtered out.
\item{{\small\tt apply-profile}} is a cognitive semantics operation that focusses on aspects of a movement event, such as source, path or goal. Here the focus is on path, which means that the event representation is annotated with a particular focus on the movement part of the trajectory and not it's goal or starting point.
\item{{\small\tt apply-dynamic-spatial-relation}} checks whether a particular spatial relation such as {\small\tt across} applies to the profiled aspect of the input movement events. It's input is a set of movement events. The output is a set of movement events that can be categorized by the spatial relation and additional information. For this, the operation queries the outcome of the spatial reasoner for the trajectories, i.e. objects, regions and their relations and tests whether it can find any trace of the spatial relation {\small\tt across} as applying to the movement event. Here it has to check whether a particular movement events trajectory includes a particular region and whether this has region has been crossed.
\end{description}

\subsection{Conceptualization strategies}
The IRL-program in Figure \ref{f:irl-program} 
is part of a particular conceptualization strategy that can
involve other dynamic spatial relations such as {\small\tt into}, {\small\tt outof} etc.
Spatial conceptualization strategies involve more than
just the choice of a spatial relation. Landmarks, perspective,
frames of reference \cite{tenbrink2007space} are all 
important aspects of the construal of spatial reality. 
All of these are implemented in our system and can be used to produce
sentences of significant complexity.

IRL includes mechanisms for the automatic and autonomous construction 
of IRL-programs. Agents use these facilities in two 
ways. First, when the speaker wants to talk about a 
particular scene, he conceptualizes an IRL-program 
for reaching that goal (see conceptualization 
in Figure \ref{f:semiotic-cycle}). Secondly, a listener trying to interpret
an utterance will construct and evaluate programs, in order, to find
the best possible interpretation of the utterance
(see interpretation in Figure \ref{f:semiotic-cycle}). 

Interpretation and conceptualization are implemented as heuristics-guided
search processes that traverse the space of possible IRL-programs.
The basic building blocks for the search are IRL-programs packaged into \emph{chunks}
which are larger structures reflecting the standard semantics of some particular
natural language, e.g. determined noun phrases in English.
The IRL search process progressively combines chunks of IRL-programs
into more and more complex IRL-programs. Each program is tested for 
compatibility with the goal of the agent, as well as the context. 

\section{Production and Interpretation of Spatial Phrases}
Robots are communicating the conceptualisation of the topic scene using an English grammar.
The grammar is implemented using a bidirectional Construction Grammar system called \emph{Fluid Construction Grammar} (FCG) \cite{steels2011design}. FCG uses one engine and a single grammar representation to support both production and interpretation. 
We implemented a spatial grammar comprising lexical items for basic concepts (e.g. block, box), events (e.g. move), as well as the spatial relations (e.g. along, across, into, out off). Additionally, we implemented a number of spatial grammar constructions. The system is similar to the one proposed in \cite{spranger2011german}. In total there are over 70 constructions.

FCG starts from a two sided feature structure. One side is for semantic features, another side for syntactic information. Constructions check if they find sufficient information to apply and subsequently if that is the case change the structure by adding information or introducing hierarchy etc. In this paper the focus is on semantics and reasoning. Nevertheless we give a short overview of the constructions involved in translating IRL-programs into spatial utterances and back. The constructions are explained here as they would apply in production. That is, given an IRL-program (e.g. Figure \ref{f:irl-program}), FCG will produce a phrase such as ``the green block moves across the red region''.
\begin{description}
\item[Lexical constructions] are bidirectional mappings between entities (bind statements in IRL-programs) and stems. For instance, there is a lexical construction for that maps {\small\tt (bind dynamic-spatial-relation ?acr across)} to the stem ``across''. Another example is the construction that maps {\small\tt (bind object-class ?block block)} to ``block'' or the construction that maps {\small\tt (bind color-category ?yll yellow)} to ``yellow''.
\item[Functional constructions] map each lexical item to a word class (also called lexical class). The idea is that the same concept can be used in different forms. For instance, a colour category can be used as an adjective such as in ``the yellow block'' or as a noun. Which one is used is determined by semantics. If the colour category is used as a  modifier such as in the operation {\small\tt apply-color} than its word class is adjective. The same is true for the spatial relation. Across is used here with the operation {\small\tt apply-dynamic-spatial-\\relation} so that the dynamic-spatial-relation preposition construction would translate it to a preposition.
\item[Phrasal constructions] take into account the larger syntactic and semantic context. An example is the adjective-noun-phrase construction, which looks for an adjective and a noun as well as a particular structure in the IRL-program and adds phrasal information such as word order. Another example is the prepositional phrasal construction that combines a preposition and a noun phrase. For the discussed example, this combines the region noun phrase and the preposition across and adds word order. Other phrasal construction handle the verb phrase and the prepositional phrase, or combine the structure into one coherent phrase.
\end{description}

Applied to an example such as the one in Figure \ref{f:irl-program} the system produces the intended sentence ``the yellow block moves across the red region''. 
\section{Discussion and Future Work}
We tested the result of the current system on the initial scenes described in this paper. The system was able to correctly produce and interpret phrases allowing robots to communicate about scenes and discriminate between scenes with different temporal and spatial characteristics. Future work will have to test the system systematically on more scenes and perform a detailed analysis.

There are a number of possible extensions and future work on the system discussed in this paper.
For example, the spatial reasoning system used in this paper has much more capabilities then employed for the purpose of this paper. The system can be used for abductive reasoning and explain missing or faulty observations \cite{dubba-bhatt-2012}. In case of the robot scenario presented in this paper, this can be used to generate or interpret descriptions where only partial information are available. E.g. consider the following scene: The object moves from the white to the yellow region, thereby passing a red region. Robot A perceives the whole narrative. It plays a description game with robot B, who is not able to see the red region because visibility is blocked. However, since it perceives the block leaving the yellow region and entering the white region, it could abduce that the red region was crossed.

Furthermore, the qualitative abstractions can be used to translate a description which is based on the point of view of one robot to the point of view of an other robot. E.g. the description provided by the speaker says: ``The green block moves from left to right.'' In order to correctly understand this description, the hearer needs be able to ``imagine'' the scene from the speakers point of view. To this end, reasoning about the robots perspective based on the position of the objects and the intrinsic-orientation of the other robot, can be used to produce or interpret viewpoint dependent descriptions of the scene. This has been studied as part of research into static locations \cite{spranger2013evolving}, but so far has not been integrated with the system presented in this paper.

In conclusion, the system presented in this paper presents a fully working system able to interpret and produce natural language phrases with dynamic spatial relations. Importantly, this is a first step towards understanding the acquisition and evolution of dynamic spatial relations. The computational reconstruction of processing will allow us to study the learning of parts of the spatial grammar, lexicon, conceptual repertoire, and ultimately setup agent-based simulations, where we can study the evolution of dynamic spatial relations.


\bibliographystyle{splncs03}


\end{document}